\def\year{2022}\relax

\documentclass[letterpaper]{article} 
\usepackage{aaai22}  
\usepackage{times}  
\usepackage{helvet}  
\usepackage{courier}  
\usepackage[hyphens]{url}  
\usepackage{graphicx} 
\usepackage{natbib}  
\usepackage{caption} 
\DeclareCaptionStyle{ruled}{labelfont=normalfont,labelsep=colon,strut=off} 
\usepackage{soul}
\usepackage{xcolor}
\usepackage[switch]{lineno}

\usepackage{amsmath}
\usepackage{tabularx}
\usepackage{amssymb}
\usepackage{arydshln}
\usepackage{booktabs}
\usepackage{subcaption}
\usepackage{multirow}
\usepackage{tikz}
\usetikzlibrary{tikzmark}

\definecolor{deepblue}{rgb}{0,0,0.5}
\definecolor{officeblue}{RGB}{0,102,204}
\definecolor{deepred}{rgb}{0.6,0,0}
\definecolor{deepgreen}{rgb}{0,0.5,0}
\definecolor{mylightgreen}{rgb}{0,128,0}
\definecolor{mybrickred}{RGB}{182,50,28}

\definecolor{fillcolor}{RGB}{216,217,252}

\definecolor{bblue}{HTML}{ADD8E6}
\definecolor{yyellow}{HTML}{FFFF00}
\definecolor{rred}{HTML}{FF6361}
\definecolor{ppink}{HTML}{BC5090}
\definecolor{ppurple}{HTML}{58508D}

\definecolor{applegreen}{rgb}{0.55, 0.71, 0.0}
\newcommand{\ModifierColor}[1]{\redbox{#1}}
\newcommand{\QuantifierColor}[1]{\bluebox{#1}}
\newcommand{\NegationColor}[1]{\greenbox{#1}}

\definecolor{olive} {cmyk} {0.64, 0.00, 0.95, 0.40}

\usepackage{tcolorbox}
\newtcbox{\redbox}{on line,
  colframe=red,colback=red!5!white,
  boxrule=0.3mm,
  arc=0.5mm,
  boxsep=1pt,left=1pt,right=1pt,top=1pt,bottom=1pt
  }

\newtcbox{\bluebox}{on line,
  colframe=blue,colback=blue!5!white,
  boxrule=0.3mm,
  arc=0.5mm,
  boxsep=1pt,left=1pt,right=1pt,top=1pt,bottom=1pt
  }

\newtcbox{\greenbox}{on line,
  colframe=applegreen,colback=applegreen!5!white,
  boxrule=0.3mm,
  arc=0.5mm,
  boxsep=1pt,left=1pt,right=1pt,top=1pt,bottom=1pt
  }
  
\newtcbox{\orangebox}{on line,
  colframe=orange,colback=orange!10!white,
  boxrule=0.3mm,
  arc=1.0mm,
  boxsep=0pt,left=1pt,right=1pt,top=1pt,bottom=1pt
  }

\newtcbox{\blueboxa}{on line,
  colframe=blue!70!black,colback=blue!5!white,
  boxrule=0.3mm,
  arc=1.0mm,
  boxsep=0pt,left=1.2pt,right=1.2pt,top=1.3pt,bottom=1.3pt
  }

\newtcbox{\blueboxb}{on line,
  colframe=blue!70!black,colback=blue!5!white,
  boxrule=0.3mm,
  arc=1.0mm,
  boxsep=0pt,left=1.2pt,right=1.2pt,top=1.6pt,bottom=1.6pt
  }

\newtcbox{\graybox}{on line,
  colframe=white,colback=gray!15!white,
  boxrule=0mm,left=1pt,right=1pt,top=1pt,bottom=1pt
  }

\makeatletter
\newcommand{\printfnsymbol}[1]{%
  \textsuperscript{\@fnsymbol{#1}}%
}
\makeatother

\begin{document}

\title{
\vspace*{-0.5in}
{{\small \hfill in Proc. of AAAI Conference on Artificial Intelligence (2022) }\\
\vspace*{.25in}} 
Probing Linguistic Information \\ For Logical Inference In Pre-trained Language Models
}

\author{Zeming Chen\thanks{Authors have equal contribution} \quad Qiyue Gao\printfnsymbol{1}\\}
\affiliations {
    Rose-Hulman Institute of Technology \\
    \texttt{\{chenz16, gaoq\}@rose-hulman.edu}
}
\maketitle
\begin{abstract}
\begin{quote}
Progress in pre-trained language models has led to a surge of impressive results on downstream tasks for natural language understanding. Recent work on probing pre-trained language models uncovered a wide range of linguistic properties encoded in their contextualized representations. However, it is unclear whether they encode semantic knowledge that is crucial to symbolic inference methods. We propose a methodology for probing linguistic information for logical inference in pre-trained language model representations. Our probing datasets cover a list of linguistic phenomena required by major symbolic inference systems. We find that (i) pre-trained language models do encode several types of linguistic information for inference, but there are also some types of information that are weakly encoded, (ii) language models can effectively learn missing linguistic information through fine-tuning. Overall, our findings provide insights into which aspects of linguistic information for logical inference do language models and their pre-training procedures capture. Moreover, we have demonstrated language models' potential as semantic and background knowledge bases for supporting symbolic inference methods. 
\end{quote}
\end{abstract}

\section{Introduction}
\label{sec:intro}
 Pre-trained language models have replaced traditional symbolic-based natural language processing systems on a variety of language understanding tasks, mainly because symbolic-based NLP systems often rely on linguistic properties as features. Those features are hard to acquire. Many types of linguistic information are either hand-written rules or background knowledge extracted from traditional knowledge base, which make symbolic-based systems hard to scale up on large benchmarks such as GLUE \cite{wang-etal-2018-glue}. On the other hand, many recent probing studies have revealed that sentence representations of pre-trained language models encode a large amount of linguistic information and background knowledge \cite{edge_probing_paper, petroni-etal-2019-language, Bouraoui2020InducingRK}. However, it remains unknown if these representations also encode implicit linguistic information for inference crucial to symbolic inference systems.

\begin{figure}[tb!]
    \centering
    \includegraphics[width=6.3cm]{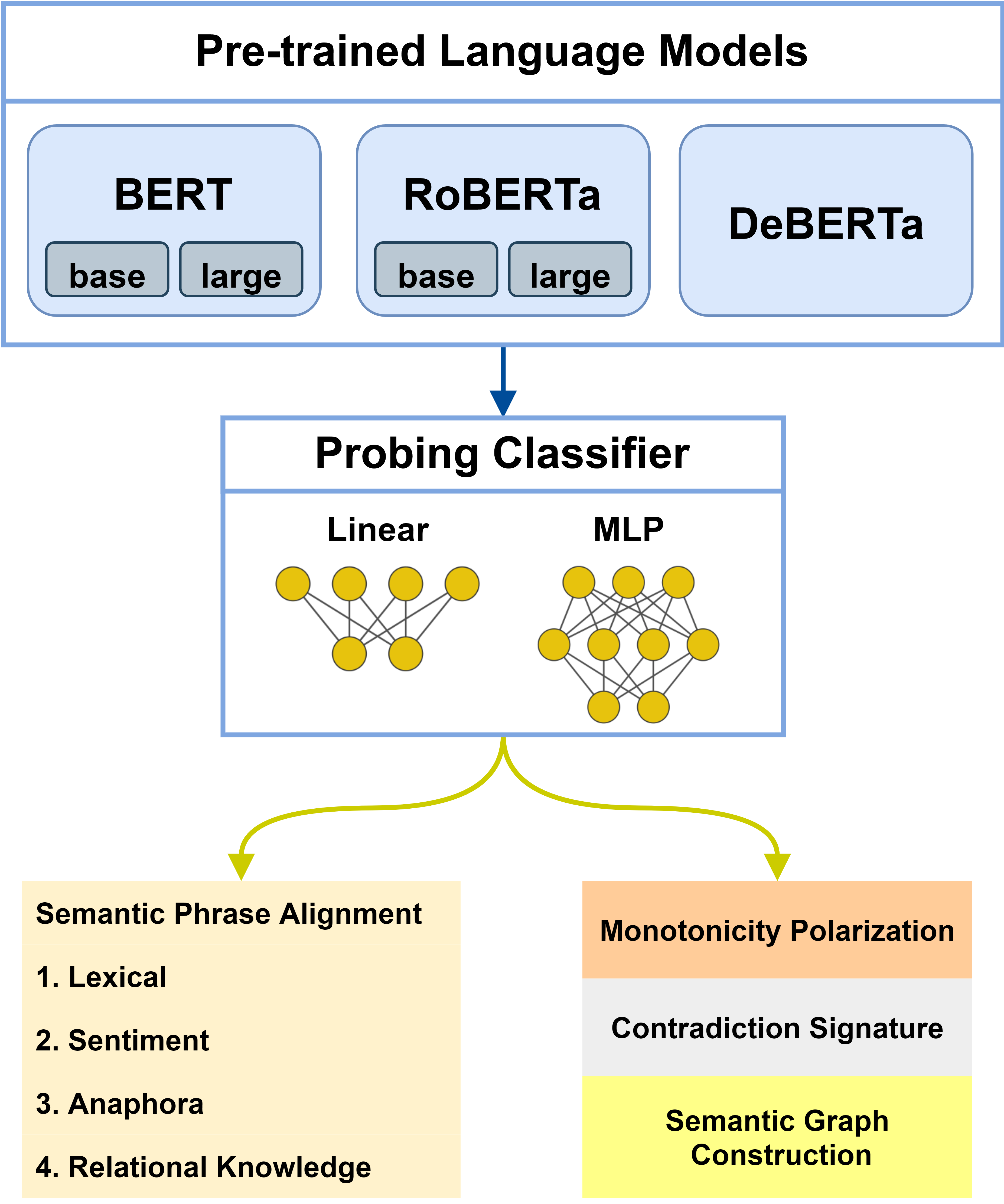}
    \caption{\small Given pre-trained language models, the probing classifier extracts linguistic infromation for a given probing task. The amount of intimation is measured by the probing accuracy and the information gain, compared with baseline word embeddings.}
    \label{fig:probe}
\end{figure}

In this paper, we propose an inference information probing framework (Figure \ref{fig:probe}). We define a set of probing tasks that focus on different types of linguistic information required by symbolic systems. In particular, we cover linguistic information for simple and complex semantic phenomena. Simple semantic phenomena often relies on partial or no context and does not require advanced linguistic skills like contextual understanding and reasoning. Our simple phenomena include word-to-word semantic relations, lexical semantics, and contradiction signatures. Complex phenomena depends on multiple types of reasoning skills like reasoning on event context, monotonicity, coreference, and commonsense knowledge. For complex phenomena, we probe sentiment, relational knowledge, anaphora resolution, and monotonicity reasoning. We are interested in answering two questions: (1) Do pre-trained language models encode linguistic information essential to symbolic inference systems? (2) Do pre-trained language models acquire new linguistic information for inference during the fine-tuning process for the NLI task? For each task, we conducted probing experiments on multiple contextualized language models and compared results to several strong baselines.

Our analysis shows that language models encode diverse types of linguistic information for inference. In particular, they encode more information on simple semantic phenomena than complex semantic phenomena. Our label-wise qualitative analysis revealed that the amount of information encoded by language models for each task is different across labels which justifies our previous findings. Moreover, we found that pre-trained language models can obtain some types of the missing linguistic information through fine-tuning for the NLI task. Overall, our findings show that pre-trained language models can be potential linguistic knowledge bases supporting symbolic inference systems. 
\paragraph{Contributions}
Our contributions are as follows: 
\begin{enumerate}
    \item Our work expands on prior probing studies by studying a wider range of linguistic information, including simple and complex semantic phenomena.
    \item Our experiments allow classifier expressiveness to be analyzed in a more complex setting covering syntactic and semantic linguistic properties beyond prior works. 
    \item Our study provides insights into what types of new linguistic information pre-trained language models obtain during fine-tuning on large NLI datasets. This contributes to the interpretability of NLI models.
\end{enumerate}
   
\section{Related Work}
Recent studies have reported the existence of linguistic properties encoded in the self-attention weights of language models' contextualized representations. These linguistic properties include syntactic structure, semantic knowledge, and some world knowledge \cite{rogers-etal-2020-primer}. Several studies train and evaluate a probing classifier on top of different language models' contextualized representations to explore the existence of information about linguistic properties. These studies have shown that pre-trained language models encode some levels of syntactic and semantic knowledge. \citet{hewitt-manning-2019-structural} recovered syntactic dependencies from BERT's embeddings by learning transformation matrices. \citet{edge_probing_paper}, which is more directly related to our work, proposed the edge probing framework and found that contextualized embeddings encode information about named entity types, relations,  semantic roles and proto roles based on the high accuracy of the probing classifier.

Some probing studies focus on inducing factual knowledge captured in pre-trained language models. A majority of the studies rely on the Masked Language Modeling (MLM) component of the model which can be adapted to induce knowledge easily, since the model only needs to fill in the blanks. \citet{petroni-etal-2019-language} showed that pre-trained BERT encodes relational knowledge competitive with those that are accessed from  knowledge bases using traditional NLP methods. They also found that BERT has strong ability to recall factual knowledge prior to any fine-tuning, making it a good candidate for open-domain QA systems (unsupervised). \citet{Bouraoui2020InducingRK} proposed a method to induce relations from pre-trained language models. They first found potential sentences that express a relation in a large corpus. A subset of sentences was used as templates. They then fine-tuned a language model to predict whether a given word pair form some relation. They found strong evidence that relations can be obtained from language models.

Compared to existing work, we extend common syntactic and semantic tasks to a range of tasks that focus on more complex linguistic phenomenons. Some of our tasks, such as semantic graph construction and monotonicity polarization, require both syntactic and semantic information. Probing for more complex linguistic tasks allows us to diagnose the particular advantages of language models over conventional NLP systems. It also allows us to study the expressiveness of probing classifiers in a more complex setting beyond syntactic tasks. Moreover, our experiments on fine-tuned NLI language models provides insights into the type of linguistic information they capture through fine-tuning.

\section{Probing Methodology}
\subsection{Edge Probing and Vertex Probing}
Edge probing is a simple and useful probing framework proposed by \citet{edge_probing_paper}. It can provide a uniform set of metrics and architectures across diverse task types. Formally, a sentence is defined as a list of tokens [$t_0, t_1, t_2, ..., t_n$] and an edge target as \{$s_1, s_2, \mathcal{L}$\} where $s_1$ and $s_2$ are two end exclusive spans with $s_1$ = \{$i^{s1}, j^{s1}$) and $s_2$ = \{$i^{s2}, j^{s2}$). $\mathcal{L}$ is a label assigned to the pair of spans which the classifier needs to accurately predict. The label set for $\mathcal{L}$ is different across tasks including both binary labels and multi labels. Each sentence [$t_0, t_1, t_2, ..., t_n$] is encoded by a language model into contextualized sentence representation [$e_0, e_1, e_2, ..., e_n$]. A projection layer concatenated with a self-attention pooling operator will be applied to the representation to extract span representations according to the index position of two spans $s_1$ = \{$i^{s1}, j^{s1}$) and $s_2$ = \{$i^{s2}, j^{s2}$). As \citet{edge_probing_paper} mentioned, the pooling is fixed-length and only operates within the bounds of a span to ensure that the classifier can only access information of the rest of the sentence from the contextual sentence representation. The two span representations are concatenated and passed to the classifier to predict the label. To ensure we only probe a pre-trained language model without modifying its parameters, we freeze its parameters to not allow for gradient updates.

The vertex probing framework has the same settings and formulations as the edge probing framework, except that vertex probing operates on every token in a sentence. The classifier receives only a single span representation as input. Formally, the definition is very similar to that of the sequence tagging task. With a list of tokens [$t_0, t_1, t_2, ..., t_n$], we define each token as a single span target $s$ = \{($i^{s}, j^{s}$), $\mathcal{L}$\}. The vertex probing is used to predict which words belong to a category in the label set. 

\begin{table*}[t!]
    \centering
    \small
    \scalebox{0.75}{\begin{tabular}{l l l}  \toprule %
    
    \textbf{Task} & \textbf{Probe Type} & \textbf{Example} \\\midrule
     
    Semantic-Graph & edge probing & A \ModifierColor{tall} \NegationColor{boy} is \QuantifierColor{running} \ModifierColor{quickly} to catch a \ModifierColor{soccer} \NegationColor{ball}. \\[5pt]
    
    (SemGraph) & & \NegationColor{[boy]} \QuantifierColor{[running]}  $\longrightarrow$ \textbf{Concept-Relation}; \ModifierColor{[tall]} \NegationColor{[boy]} $\longrightarrow$ \textbf{Concept-Modifier};\\[5pt] 
    
    & & \ModifierColor{[quickly]} \QuantifierColor{[running]} $\longrightarrow$ \textbf{Relation-Modifier} \\[5pt] \midrule
    
    Monotonicity & vertex probing & Some$^\uparrow$ people$^\uparrow$ in$^\uparrow$ the$^\uparrow$ White$^=$ House$^=$ does$^\uparrow$ not$^\uparrow$ know$^\downarrow$ if$^\downarrow$ any$^\downarrow$ dog$^\downarrow$ in$^\downarrow$ Ohio$^\downarrow$ ate$^\downarrow$ bananas$^\downarrow$ yesterday$^\downarrow$ \\[5pt] \midrule
    
    Lexical & edge probing & P: The \QuantifierColor{[man]$_{s^1}$} is holding a \ModifierColor{[saxophone]$_{s^2}$} \\[5pt]
    (SA-Lex) & & H: The man is holding an \ModifierColor{[instrument]$_{s^3}$}. \\[5pt]
    & & ($s_1$, $s_3$) $\longrightarrow$ \textbf{Unaligned}; ($s_2$, $s_3$) $\longrightarrow$ \textbf{Aligned} \\\midrule
     
    Anaphora & edge probing & The \QuantifierColor{[technician]$_{s^1}$} told the \ModifierColor{[customer]$_{s^2}$} that \ModifierColor{[he]$_{s^3}$} could pay with cash.  \\[5pt]
    (SA-AP) & & ($s_1$, $s_3$) $\longrightarrow$ \textbf{Unaligned}; ($s_2$, $s_3$) $\longrightarrow$ \textbf{Aligned} \\\midrule
    
    Sentiment & vertex probing   & P: When asked about the restaurant, Brielle said, ``\ModifierColor{[It]$_{t^1}$ [was]$_{t^2}$ [terrible!]$_{t^3}$} \\
    (SA-ST) & & \QuantifierColor{[I]$_{t^4}$ [found]$_{t^5}$ [this]$_{t^6}$ [product]$_{t^7}$ [to]$_{t^8}$ [be]$_{t^9}$ [way]$_{t^{10}}$ [too]$_{t^{11}}$ [big]$_{t^{12}}$} '' \\[5pt]
    & & H: Brielle \ModifierColor{[did]$_{t^{13}}$ [not]$_{t^{14}}$ [like]$_{t^{15}}$ [the]$_{t^{16}}$ [restaurant]$_{t^{17}}$}  \\[5pt]
    & & \{$t^1$, $\dots$, $t^3$\} $\longrightarrow$ \textbf{Align}$_1$; \{$t^4$, $\dots$, $t^{12}$\} $\longrightarrow$ \textbf{Unaligned}; \{$t^{13}$, $\dots$, $t^{17}$\} $\longrightarrow$ \textbf{Align}$_2$ \\\midrule
    
    Relational-Knowledge & vertex probing & P: \ModifierColor{[Dirk]$_{t^1}$ [Nowitski]$_{t^2}$ [is]$_{t^3}$ [a]$_{t^4}$ [current]$_{t^5}$ [NBA]$_{t^6}$ [star]$_{t^7}$} playing with the \\[5pt]
    (SA-RK) & & \;\;\; \QuantifierColor{[Dallas]$_{t^8}$ [Mavericks]$_{t^9}$ [as]$_{t^{10}}$ [an]$_{t^{11}}$ [all-purpose]$_{t^{12}}$ [forward]$_{t^{13}}$} \\[5pt]
    & & H: \ModifierColor{[Dirk]$_t^{14}$ [Nowitski]$_t^{15}$ [plays]$_t^{16}$ [in]$_t^{17}$ [the]$_t^{18}$ [NBA]$_t^{19}$} \\[5pt]
    & & \{$t^1$, $\dots$, $t^7$\} $\longrightarrow$ \textbf{Align}$_1$; \{$t^8$, $\dots$, $t^{13}$\} $\longrightarrow$ \textbf{Unaligned}; \{$t^{14}$, $\dots$, $t^{19}$\} $\longrightarrow$ \textbf{Align}$_2$ \\\midrule
    
    Contradiction Signature & vertex probing & P: Italy and Germany \QuantifierColor{[have]$_{t^1}$ [each]$_{t^2}$ [played]$_{t^3}$ [twice]$_{t^4}$}, \\[5pt]
     (ContraSig) & & \;\;\;  and they \NegationColor{[haven't]$_{t^5}$ [beaten]$_{t^6}$ [anybody]$_{t^7}$ [yet]$_{t^8}$} \\[5pt]          
     & & H: Italy \NegationColor{[defeats]$_{t^9}$ [Germany]$_{t^{10}}$} \\[5pt]
     & & \{$t^1$, $\dots$, $t^4$\} $\longrightarrow$ \textbf{None}$_1$; \{$t^5$, $\dots$, $t^8$\} $\longrightarrow$ \textbf{Contra-sig}$_1$; \{$t^9$, $t^{10}$\} $\longrightarrow$ \textbf{Contra-sig}$_2$ \\
      
      \bottomrule
      \end{tabular}}
    \caption{\small This table lists examples of our probing tasks. For semantic graph construction task, the red, green and blue boxes denote modifiers, concepts and relations respectively. For semantic alignment tasks and the contradiction signature detection task, the red boxes are spans that are semantically aligned. The green boxes are spans that form a contradiction signature. The blue boxes are spans that are irrelevant to semantic alignment or contradiction. Here P stands for a premise, and H stands for a hypothesis. The anaphora-based alignment only uses a single sentence. For the labels, ($s^1, s^3$) $\longrightarrow$ \textbf{Aligned} means that $s^1$ and $s^3$ are aligned. \{$t^1$, $\dots$, $t^7$\} $\longrightarrow$ \textbf{Align}$_1$ means that tokens $t^1$ to $t^7$ belongs to the first phrase in a semantically aligned pair.}
    \label{tab:data_exp}
\end{table*}

\subsection{Classifier Selection}
Selecting a good probing classifier is essential to the probing process. We first choose the linear classifier. According to \citet{hewitt-liang-2019-designing}, the linear classifier is less expressive and thus is prevented from memorizing the task. However, \citet{pimentel-etal-2020-information} uses probing to estimate the mutual information between a representation-valued and a linguistic-property–valued random variable. They argue that the most optimal probe should be used to minimize the chance of misinterpreting a representation's encoded information, and therefore achieve the optimal estimate of mutual information. 
To lessen the chance of misinterpretation, we conducted probing using a Multi-layer Perceptron (MLP) classifier with one hidden layer.  

\subsection{Experiment Setup} To answer both questions 1 and 2 in the introduction, we experiment with five pre-trained language models. We selected BERT-base and BERT-large \cite{devlin-etal-2019-bert}, RoBERTa-base and RoBERTa-large \cite{Liu2019RoBERTaAR}, and DeBERTa \cite{he2021deberta}. All five models can provide contextualized sentence representations and have shown impressive performance on the GLUE \cite{wang-etal-2018-glue} benchmark. Our experiment setup follows three types of evaluation methods to interpret the probing performance.

\paragraph{Probing Accuracy} We probe pre-trained language models and the baseline word embeddings using linear and MLP classifiers. Then, we compare their performance to determine if the pre-trained models improve over the baselines. If such improvement is significant, the pre-trained models contain more information about a task than the baseline. Otherwise, they do not contain enough information to benefit a task. We select four uncontextualized word embeddings as our baselines, including random embedding, FastText \cite{joulin2017bag}, Glove \cite{pennington-etal-2014-glove}, and Word2Vec \cite{word2vec2013}. We also conduct a label-wise qualitative analysis for each task to explore if the amount of information is encoded differently across the task-specific labels. Finally, to determine if language models can learn the missing linguistic information for inference, we evaluate NLI models fine-tuned on MultiNLI \cite{williams-etal-2018-broad}, using probing tasks that do not benefit from the pre-trained models. 

\paragraph{Control Task} \citet{hewitt-liang-2019-designing} argue that accuracy cannot fully validate that a representation encodes linguistic information since a highly expressive classifier could have memorized the task. They proposed the use of control tasks to complement probings. The main idea is to validate if a classifier could predict task outputs independently of a representation's linguistic properties. If the classifier can achieve high accuracy in this setting, the accuracy does not necessarily reflect the properties of the representation. A control task has the same input as the associated linguistic task, but it assigns random labels to the input data. The selectivity, which is the difference between the linguistic task accuracy and the control task accuracy, is used to measure the quality of a probe. A good probe should have high selectivity meaning that it has low control task accuracy but high linguistic task accuracy.

\paragraph{Information-theoretic Probing} \citet{pimentel-etal-2020-information} argue that the task of supervised probing is an attempt to measure how much information a neural representation can provide for a task. They operationalized probing as estimating the mutual information between a representation and a probing task. The mutual information from a target representation is compared to the information estimated from a control function’s representation, which serves as a baseline for comparison. In our experiments, we use uncontextualized baselines as control functions. We estimate the information gain between a contextualized embedding and a baseline. Information gain measures how much more information about a task a contextualized embedding has over a baseline. In addition, we transform the gain into a percentage measurement to make the results more interpretable.

\section{Inference Information Probes}
In this section, we introduce a list of edge and vertex probing tasks for probing implicit linguistic information for symbolic inference methods in pre-trained language model representations. To discover potential tasks that can provide essential linguistic information for symbolic inferences, we studied four major logical systems for NLI, all with high accuracy on SICK \cite{Marelli2014ASC}, and several challenge datasets for the NLI task. They include NLI systems based on natural logic \cite{abzianidze-2020-learning}, monotonicity reasoning \cite{hu-etal-2020-monalog, chen-etal-2021-neurallog}, and theorem proving \cite{yanaka-etal-2018-acquisition}.

\subsection{Semantic Graph Construction (SemGraph)}
This task probes the graph-based abstract meaning representation for sentences, a type of knowledge found effective in symbolic systems for acquiring paraphrase pairs and selecting correct inference steps \cite{yanaka-etal-2018-acquisition, chen-etal-2021-neurallog}. The task is to construct a semantic graph that captures connections between concepts, modifiers, and relations in a sentence. Relations are words that form a connection, including verbs and prepositions. Concepts are arguments connected by a relation such as objects and subjects. Each concept connects to a set of modifiers that attribute to it. An example semantic graph is shown in Table \ref{tab:data_exp}. We define this as an edge probing task and assign a label to a pair of tokens. A label is selected from the label set: concept-to-relation, concept-to-modifier, relation-to-concept, relation-to-modifier, relation-to-relation, modifier-to-relation, modifier-to-concept. To construct the dataset, we use dependency parsing and semantic role labeling tools to identify concepts, modifiers, and relations in a sentence and the connection between them. We selected premises from the SNLI test set as our inputs and split them into training and testing sets.

\begin{table}[t!]
    \centering
    \footnotesize
    \scalebox{0.80}{
    \begin{tabular}{l l r r}
        \toprule
        \textbf{Task} & \textbf{Split} & \textbf{\# S} & \textbf{Origin}  \\\midrule
        
        \multirow{2}{*}{\textbf{SemGraph}}
        & train & 10,000 & \cite{bowman-etal-2015-large}        \\     
        & test & 5,000   & \cite{bowman-etal-2015-large}         \\ \hline
        
        \multirow{2}{*}{\textbf{ContraSig}}
        & train & 1,000 & \cite{Marelli2014ASC}                 \\     
        & test & 500    & \cite{Marelli2014ASC}           \\ \hline
        
        \multirow{2}{*}{\textbf{Monotonicity}}
        & train & 5,000 & \cite{yanaka-etal-2019-neural}        \\     
        & test & 500    & \cite{chen-gao-2021-IWCS}               \\ \hline
        
        \multirow{2}{*}{\textbf{SA-Lex}}
        & train & 1,000 & \cite{glockner-etal-2018-breaking}    \\     
        & test & 500    & \cite{glockner-etal-2018-breaking}      \\ \hline
        
        \multirow{2}{*}{\textbf{SA-ST}}
        & train & 1,000 & \cite{poliak-etal-2018-collecting}    \\     
        & test & 600    & \cite{poliak-etal-2018-collecting}      \\ \hline
        
        \multirow{2}{*}{\textbf{SA-AP}}
        & train & 500 & \cite{rudinger-etal-2018-gender}        \\     
        & test & 220  & \cite{rudinger-etal-2018-gender}        \\ \hline
        
        \multirow{2}{*}{\textbf{SA-RK}}
        & train & 1,000 & \cite{poliak-etal-2018-collecting}    \\     
        & test & 500    & \cite{poliak-etal-2018-collecting}      \\
        
        \bottomrule
    \end{tabular}}
    
    \caption{ \small This table shows dataset details for each probing task. We list the number of training and testing examples, and also the datasets that the train and test sets are recast from.}
    \label{tab:data_detail}
\end{table}

\subsection{Semantic Alignment (SA)}
This set of tasks probes the linguistic information for inference involving semantically aligned phrase or word pairs. These aligned pairs can often serve as an explanation of the entailment gold-label \cite{abzianidze-2020-learning, chen-etal-2021-neurallog}. 
We cover a wide range of semantic phenomena common in natural language inference including lexical (SA-Lex), anaphora (SA-AP), sentiment (SA-ST), and relational knowledge (SA-RK). Table \ref{tab:data_exp} lists each type of semantic alignment with associated examples. Probing data are first collected from multiple challenge datasets for NLU and then are manually annotated for the edge and vertex probing framework. For the sentiment task, we noticed that the aligned phrases are always part of a person's saying, leading a model to solve the task quickly by memorization. To avoid this, we concatenated each premise with speech fragments from another randomly selected premise to build a more complex premise. For instance, in the example in Table \ref{tab:data_exp}, \textit{I found this product to be way too big} is a speech fragment from another premise sample.


\begin{table*}[t!]
    \centering
    \footnotesize
    \scalebox{0.85}{
    \begin{tabular}{l c c c c c c c c}
        \toprule
            \textbf{Model} & \textbf{SemGraph} & \textbf{ContraSig} & \textbf{Monotonicity} & \textbf{SA-Lex} & \textbf{SA-ST} & \textbf{SA-AP} & \textbf{SA-RK} \\
        \hline 
        
        \multicolumn{9}{c}{Group 1: \textbf{Baselines} (direct probing, no fine-tuning)} \\ \hline 
        \textbf{Random}     & 68.5 & 29.7 & 48.8 & 45.5 & 46.8 & 48.9 & 45.7 \\
        \textbf{Word2Vec}   & 68.8 & 42.4 & 43.4 & 59.7 & 31.4 & 32.3 & 42.3 \\
        \textbf{Glove}      & 71.3 & 40.2 & 41.3 & 60.6 & 33.4 & 35.8 & 50.5 \\
        \textbf{FastText}   & 69.4 & 31.7 & 51.9 & 50.1 & 51.3 & 51.3 & 52.2 \\
        \hline
 
        \multicolumn{9}{c}{Group 2: \textbf{Language Models} (\textbf{Linear} classifier probing, no fine-tuning; \textbf{selectivity} is shown in parenthesis)} \\ \hline
        \textbf{BERT}-base      & 91.8 (42.0) & 58.5 (35.1) & 48.5 (35.8) & 57.4 (7.30) &  56.3 (34.7) & 49.8 (0.1) &  62.8 (50.3) \\
        \textbf{BERT}-large     & 88.9 (39.5) & 51.6 (37.9) & 44.9 (37.4) & 55.5 (4.80) &  51.2 (33.1) & 50.2 (0.1) &  62.1 (49.5) \\
        \cdashline{1-9}
        \textbf{RoBERTa}-base   & 88.9 (39.2) & 52.2 (33.0) & 42.8 (42.3) & 61.4 (11.3) &  47.5 (38.6) & 49.4 (0.6) & 62.3 (49.7) \\
        \textbf{RoBERTa}-large  & 89.6 (40.0) & 48.4 (27.6) & 38.1 (43.0) & 66.2 (16.2) &  46.1 (36.6) & 49.8 (0.3) & 62.9 (50.1) \\
        \cdashline{1-9}
        \textbf{DeBERTa}        & \textbf{93.4} (43.7) & 78.5 (32.3) & 54.9 (42.2) & 83.8 (33.2) &  42.8 (35.7) & 51.8 (2.3) & 65.6 (52.8) \\
        \hline
        
        \multicolumn{9}{c}{Group 3: \textbf{Language Models} (\textbf{MLP} classifier probing, no fine-tuning; \textbf{selectivity} is shown in parenthesis)} \\ \hline
        \textbf{BERT}-base      & 90.7 (40.7) & 91.6 (31.7) & 58.1 (41.3) & 89.0 (38.8) &  67.1 (34.3) & 61.5 (17.6) & 70.9 (54.7) \\
        \textbf{BERT}-large     & 89.3 (38.0) & 91.0 (33.0) & 57.4 (34.6) & 88.8 (38.4) &  \textbf{67.4} (35.5) & 77.6 (31.3) & 69.0 (53.1) \\
        \cdashline{1-9}
        \textbf{RoBERTa}-base   & 91.5 (39.5) & 92.5 (27.8) & 50.1 (49.1) & 87.2 (36.6) &  66.1 (37.3) & 87.0 (38.6) & 70.9 (52.9) \\
        \textbf{RoBERTa}-large  & 90.1 (40.6) & 91.9 (35.5) & 55.4 (46.8) & 88.9 (37.2) &  65.7 (38.2) & \textbf{88.6} (39.3) & 70.2 (49.1) \\
        \cdashline{1-9}
        \textbf{DeBERTa}        & 92.3 (42.2) & \textbf{92.9} (32.5) & \textbf{65.7} (58.6) & \textbf{91.0} (41.5) &  63.0 (36.9) & 85.2 (37.3) & \textbf{72.9} (58.2) \\
        
        \bottomrule 
    \end{tabular}}
    
    \caption{\small This table lists results from the probing and fine-tuning experiments We include in group 1 probing accuracy (\%) from two baselines. Group 2 and 3 shows probing accuracy (\%) of the linear classifier (group 2) and the MLP classifier (group 3). For each probe, we also record the selectivity score (\%) from control tasks.}
    \label{tab:instance}
\end{table*}

We formulate each task as either an edge probing or a vertex probing task during annotation. For edge probing tasks, we assign either \textbf{Aligned} or \textbf{Unaligned} to a pair of spans. For example, in the Lexical example in Table \ref{tab:data_exp}, ($s^2$: [\textit{saxophone}], $s^3$: [\textit{instrument}]) are aligned, and ($s^1$: [\textit{man}], $s^3$: [\textit{instrument}]) are unaligned. In a vertex probing task, we label a token as either \textbf{Aligned$_1$} (the token belongs to the first phrase of the aligned pair), \textbf{Aligned$_2$} (the token belongs to the second phrase of the aligned pair), or \textbf{Unaligned} (the token is not in any aligned phrases). For example, in the relational-knowledge example in Table \ref{tab:data_exp}, \{\textit{Dirk}, \textit{Nowitski}, \textit{is}, \textit{a}, \textit{current}, \textit{NBA}, \textit{star}\} are tokens in the first phrase of the aligned pair, \{\textit{Dirk}, \textit{Nowitski}, \textit{plays}, \textit{in}, \textit{the}, \textit{NBA}\} are tokens in the second phrase of the aligned pair, and \{\textit{Dallas}, \textit{Mavericks}, \textit{as}, \textit{an}, \textit{all-purpose}, \textit{forward}\} are unaligned tokens. We apply edge probing to tasks with single word spans (lexical and anaphora) and vertex probing to tasks involving multi-word spans (sentiment and relational knowledge). In general, vertex probing adds more distractors into the task to increase the difficulty level, ensuring that the models are using the correct types of reasoning when making a decision. 

\subsection{Contradiction Signature (ContraSig)}
Being able to reason about contradictions between a pair of sentences is a fundamental requirement of Natural Language Inference. To determine a contradictory relationship, systems often rely on contradiction signatures, or possible rationales of contradiction in the sentence pair. Contradiction signature detection tests for both syntax and fundamental semantics. We define this task as vertex probing and manually annotated one dataset for detecting contradiction in text \cite{Marelli2014ASC} by labeling tokens in the first phrase of a contradiction signature as \textbf{Contra-sig$_1$}, tokens in the second phrase of a contradiction signature as \textbf{Contra-sig$_2$}, and irrelevant tokens as \textbf{None}. Table \ref{tab:data_exp} shows an example, with \{\textit{have}, \textit{each}, \textit{played}, \textit{twice}\} as irrelevant tokens, \{\textit{defeats}, \textit{Germany}\} as tokens in the first phrase of the contradiction signature, and \{\textit{haven't}, \textit{beaten}, \textit{anybody}, \textit{yet}\} as tokens in the second phrase of the contradiction signature.

\subsection{Monotonicity Polarization (Monotonicity)}
Monotonicity information supports word-replacement-based logical inferences that NLI systems can use. For each token, we assign a monotonicity mark that is either \textbf{Monotone}  ($\uparrow$), \textbf{Antitone} ($\downarrow$), or \textbf{None} ($=$). To construct our dataset, we annotated monotonicity information on all sentences in the MED dataset \cite{yanaka-etal-2019-neural} as training examples using a monotonicity annotation tool called Udep2Mono \cite{chen-gao-2021-IWCS}. For testing, we extended a challenging gold-label monotonicity dataset used by Udep2Mono that includes multiple levels of monotonicity changes from different quantifiers and logical operators. For each sentence, we replicate ten sentences following the same syntactic format.  Vertex probing is used for monotonicity polarization because a model must predict every token's monotonicity information.

\section{Experiment Results and Findings}
\subsection{Do LMs encode information for inference?}
Here we evaluate the degree to which pre-trained language models encode implicit information of linguistic properties that is essential to logical inference systems. We conducted probes on the five pre-trained language models. Table \ref{tab:instance} shows results from the probing experiments.

\paragraph{Semantic Graph Construction}
With the linear classifier, all language models achieve high probing accuracy that outperforms the baselines. Together, with high selectivity, this is strong evidence that the information in these models' representations allows the classifier to recover the connectives of concepts, relations, and modifiers. Interestingly, the MLP classifier did not improve on the linear classifier significantly, suggesting that the information is easy to interpret. The performance here is consistent with language models' good performance on dependency parsing, semantic role labeling, and part-of-speech tagging \cite{edge_probing_paper} which are related to semantic graph construction.

\paragraph{Semantic Alignment}
We observe that on lexical and anaphora based alignments (SA-Lex, SA-AP), language models show high probing accuracy that improves over the baselines significantly when using the MLP Classifier. This is evidence that language models encode linguistic information of these two types of semantic alignment since they also show high selectivity. Language models do not improve over the baselines when using the linear classifier, suggesting that these types of linguistic information might be hard to interpret. Language models only improved trivially over the baselines for alignments involving sentiment and relational knowledge (SA-ST, SA-RK). The insignificant improvement suggests that models weakly capture the information on these complex semantic alignment. Overall, the trend is that pre-trained models tend to show poor performance on complex semantic alignment that requires understanding and reasoning. This behavior is consistent with \citet{edge_probing_paper}'s finding that language models only encode a small amount of information on semantic reasoning.

\paragraph{Contradiction Signature and Monotonicity}
For the task on contradiction signature detection, all language models show poor performance with linear classifier except DeBERTa, which has relatively high accuracy (78.5\%), validating that information on contradiction signatures is more accessible from DeBERTa than the other four models. After using the MLP classifier, all models' accuracy increased significantly (above 90\%) while maintaining very high selectivity. We attribute this partly to the fact that many contradictions are simple morphological negation and antonyms, which can be largely detected by using lexical semantics and syntax. The high accuracy is thus strong evidence that language models do encode a good amount of information on syntax and lexical semantics. For the monotonicity polarization task, language models show low accuracy with both the linear classifier and the MLP classifier. This suggests that these language models may not encode much monotonicity information that can support the polarization. Again, the results here show that pre-trained models encode more information on simple semantic phenomena (contradiction signature) than complex ones (monotonicity).

\subsection{Label-wise Qualitative Analysis}
\begin{figure}[t!]
    \centering
    \includegraphics[width=7cm]{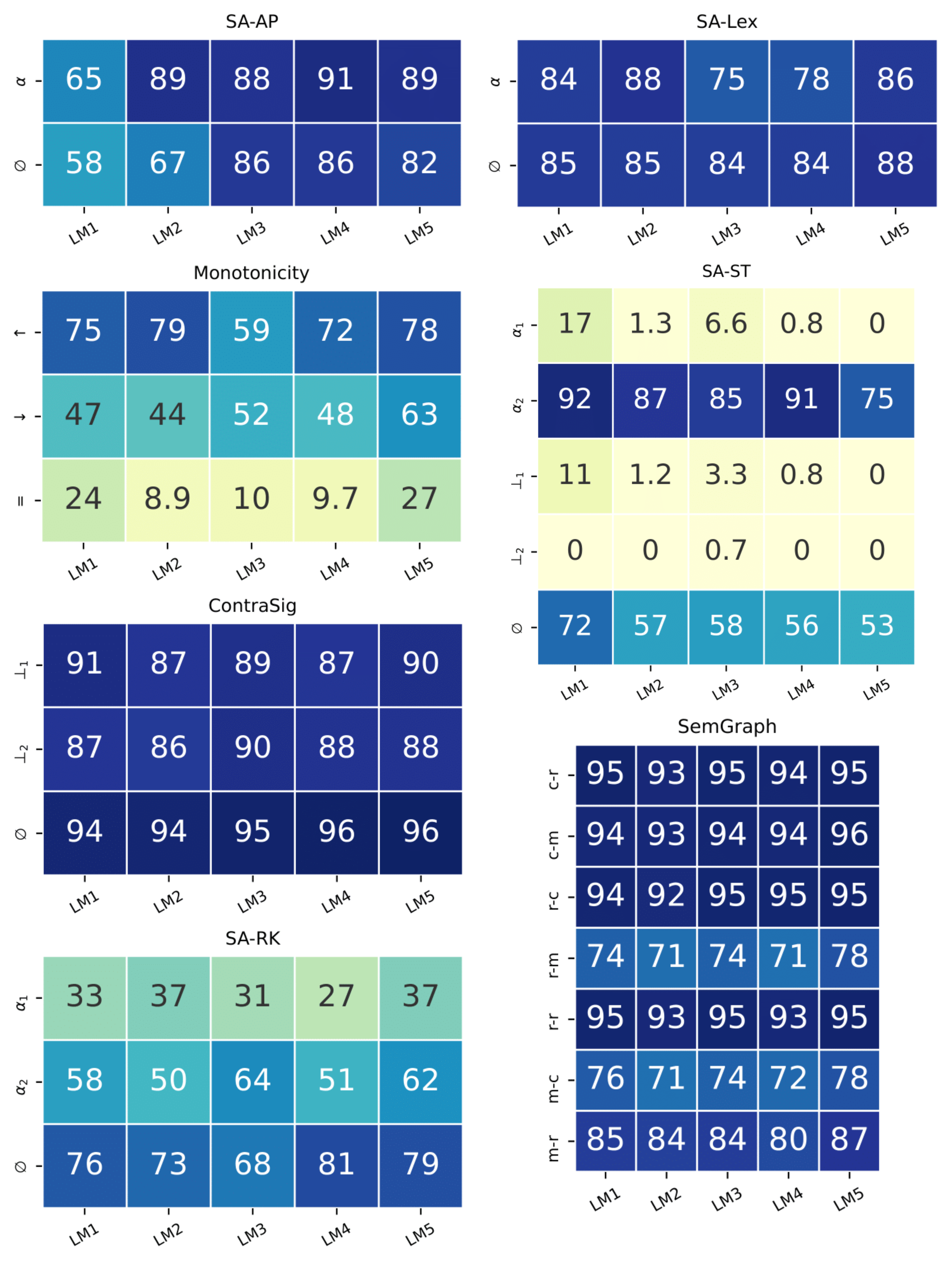}
     \caption{\small Plots here shows label-wise accuracy across models for each inference information probing task. Here LM1-5 stands for the five language models in order (BERT-base, BERT-large, RoBERTa-base, RoBERTa-large, DeBERTa).}
     \label{fig:acc_label}
\end{figure}

To further understand the amount of linguistic information that pre-trained language models capture, we analyze label-wise probing quality for each task. Label-wise accuracy per task are shown in figure \ref{fig:acc_label}.
We first observe that on some tasks (SA-Lex, SA-AP, ContraSig, SemGraph), pre-trained language models show high and balanced accuracy across labels. These behaviors are strong evidence that these models encode rich information on simple semantic phenomena. On the semantic graph construction task, the accuracy distribution is similar across models. The relatively low accuracy on modifier-to-relation (m-r) and modifier-to-concept (m-c) show the incompleteness of information in language models that support the linking of modifiers to words being modified. Language models seem to encode information for linking concepts to corresponding words since the accuracy is consistently high. Note that although anaphora (SA-AP) is a complex phenomena, models show decent performance which validates the finding from the previous experiment. 

For other complex semantic alignment (SA-ST, SA-RK), the heatmaps show highly imbalanced label-wise accuracy. The models have higher accuracy on words in the hypothesis of an aligned pair than words in the premise, which has a much more complicated context. Since vertex-probing needs to locate phrases in a premise contributing to an entailment, the low accuracy on predicting span locations in a premise suggests that language models only encode very little linguistic information on complex semantic phenomena. For monotonicity polarization, the accuracy on each label is very different. Across language models, the accuracy on monotone polarity is higher than that on antitone and neutral polarity. This is consistent with other findings from other probing studies on monotonicity reasoning. \cite{yanaka-etal-2019-neural, geiger-etal-2020-neural}. The results from this analysis validates our main finding from the previous experiment: models tend to encode less information on complex semantic phenomena. 

\subsection{Discussions}
\paragraph{Information-Theoretic Probing}
We conducted additional experiments on information-theoretical probing to validate our findings based on probing accuracy. Recall that here we want to estimate the information gain between a pre-trained representation and a baseline representation. We followed \citet{pimentel-etal-2020-information}'s practice by using a more powerful probing classifier (MLP). We compared each language model to the best-performed baseline embedding. As table \ref{tab:information_gain} shows, pre-trained language models on average encode more than 50\% more information on the eight probing tasks. Overall, pre-trained language models encode more information than baseline embedding consistently across all tasks. Among these, the highest information gain of language models is on lexical alignment (more than 100\% of increase). This is surprising since baseline word embeddings are a representation of word semantics. We hypothesize that this is due to the proximity of words that contradicts each other in the embedding space. Based on the results, we conclude that pre-trained language models encode significantly more information on linguistic information of logical inference than conventional word embeddings. 

\paragraph{Classifier Expressiveness} Some of our findings contradict several statements made by \citet{hewitt-liang-2019-designing} regarding classifier expressiveness. First, they claim that one should choose a less expressive classifier over a highly expressive one (linear over MLP) since the prior one has higher selectivity. However, based on the accuracy, we observe that the linear classifier has worse performance for semantically-oriented tasks than the MLP classifier. We also found that the MLP classifier can achieve similar selectivity as a linear classifier for these tasks while achieving higher accuracy. These findings suggest that linear classifiers could misinterpret semantic information from a representation, which supports \citet{pimentel-etal-2020-information}'s arguments on using a more powerful classifier. Secondly, they claim that a probing classifier with sufficient expressiveness can learn any task on top of a lossless representation with enough training examples. However, we found that even high expressive classifiers like MLP fail to perform well on tasks with monotonicity and higher-level semantic reasoning.


\begin{table}[t!]
    \centering
    \footnotesize
    \scalebox{0.75}{
    \begin{tabular}{l l l l l}
    
    \toprule
    \textbf{Model} & \textbf{SemGraph} & \textbf{ContraSig} & \textbf{Monotonicity} & \textbf{SA-Lex} \\ \hline 
 
        \textbf{BERT}-base      
        & 0.10 (6\%) & 1.16 (65\%) & 0.82 (58\%) & 1.16 (116\%) \\
        
        \textbf{BERT}-large     
        & 0.08 (5\%) & 1.15 (64\%) & 0.8 (56\%) & 1.14 (114\%) \\
        
        \cdashline{1-5}
        
        \textbf{RoBERTa}-base   
        & 0.10 (6\%) & 1.16 (65\%) & 0.85 (60\%) & 1.14 (114\%) \\
        
        \textbf{RoBERTa}-large  
        & 0.09 (5\%) & 1.14 (64\%) & 0.9 (63\%) & 1.15 (115\%) \\
        
        \cdashline{1-5}
        
        \textbf{DeBERTa}       
        & 0.11 (7\%) & 1.16 (65\%) & 0.78 (55\%) & 1.16 (116\%) \\ 
        
        \hline
    
    & \textbf{SA-ST} & \textbf{SA-AP} & \textbf{SA-RK} & \textbf{Avg} \\ \hline
    
        \textbf{BERT}-base      
        & 0.39 (39\%) & 1.07 (54\%) & 0.74 (80\%) & 0.78 (59\%) \\
        
        \textbf{BERT}-large     
        & 0.41 (41\%) & 1.02 (51\%) & 0.74 (80\%) & 0.76 (59\%) \\
        
        \cdashline{1-5}
        
        \textbf{RoBERTa}-base   
        & 0.74 (74\%) & 1.04 (52\%) & 0.80 (86\%) & 0.83 (65\%) \\
        
        \textbf{RoBERTa}-large  
        & 0.62 (62\%) & 1.05 (53\%) & 0.78 (84\%) & 0.82 (64\%) \\
        
        \cdashline{1-5}
        
        \textbf{DeBERTa}       
        & 0.67 (67\%) & 1.02 (51\%) & 0.67 (72\%) & 0.79 (62\%) \\ 

    \bottomrule 
    \end{tabular}}
    
    \caption{\small Results on infromation-theoretic probing. Here we show the amount of extra information each language model shares, compared to the best baseline word embedding. The percentage of information gain is listed in the parenthesis.}
    \label{tab:information_gain}
\end{table}

\begin{table}[t!]
    \centering
    \footnotesize
    \scalebox{0.8}{
    \begin{tabular}{l c c c}
        \toprule
            \textbf{Model} & \textbf{Monotonicity} &\textbf{SA-ST} & \textbf{SA-RK} \\
        \hline 
        
        \textbf{BERT}-base  & 60.9 ({\large \color{cyan} $\blacktriangle$} 2.8) & 79.8 ({\large \color{cyan} $\blacktriangle$} 12.7) & 81.4 ({\large \color{cyan} $\blacktriangle$} 14.2) \\
       
        \textbf{BERT}-large & 60.2 ({\large \color{cyan} $\blacktriangle$} 2.8) & 77.6 ({\large \color{cyan} $\blacktriangle$} 10.2) & 80.5 ({\large \color{cyan} $\blacktriangle$} 14.6) \\
        
        \cdashline{1-4}
        
        \textbf{RoBERTa}-base   
        & 59.4 ({\large \color{cyan} $\blacktriangle$} 9.3) & 78.0 ({\large \color{cyan} $\blacktriangle$} 11.9) & 88.0 ({\large \color{cyan} $\blacktriangle$} 22.7) \\
        
        \textbf{RoBERTa}-large 
        & 58.0 ({\large \color{cyan} $\blacktriangle$} 2.6) & 76.1 ({\large \color{cyan} $\blacktriangle$} 10.4) & 89.7 ({\large \color{cyan} $\blacktriangle$} 27.7) \\
        
        \cdashline{1-4}
        
        \textbf{DeBERTa}        
        & 68.7 ({\large \color{cyan} $\blacktriangle$} 3.0) & 78.9 ({\large \color{cyan} $\blacktriangle$} 14.9) & 89.6 ({\large \color{cyan} $\blacktriangle$} 19.1) \\
        
        \bottomrule 
    \end{tabular}}
    
    \caption{\small Probing accuracy (\%) of fine-tuned NLI models on probing tasks that did not benefit from pre-trained models.}
    \label{tab:finetune}
\end{table}

\paragraph{Can LMs learn missing information?}
Here we evaluate whether pre-trained language models can obtain linguistic information for inference, missing from their pre-trained representations, through fine-tuning for the NLI task. We select a version fine-tuned on the MultiNLI dataset for each language model. We probed these fine-tuned models for three tasks that did not benefit from the pre-trained language models. Our probing results are shown in Table \ref{tab:finetune}, and we only record probings with the best performance here. We observe that all language models' fine-tuned representations improve over their pre-trained representations significantly on the sentiment (SA-ST) and relational-knowledge-based (SA-RK) semantic alignment. In contrast, they do not improve the performance on monotonicity polarization (Monotonicity). The results show that language models can capture linguistic information on some types of semantic reasoning but not on monotonicity. Possible explanations are that these models could not obtain monotonicity information during fine-tuning, or the training data of MultiNLI does not contain enough examples about monotonicity.  


\section{Conclusions and Future Work}
We presented a systematic study on determining if pre-trained language models encode implicit linguistic information essential to symbolic inference methods. For each probing task, we constructed associating datasets. We then conducted probings on pre-trained language models using both linear and MLP classifiers for each task. In general, we first found that baseline word embeddings do not contain much linguistic information for inference and contextualized representations of language models encode some levels of linguistic information. However, they encode more information on simple semantic phenomena than on complex phenomena. Moreover, we found that linear classifiers can predict correctly under syntactical information but often misinterpret information on semantics resulting in low classification accuracy. Our label-wise qualitative analysis found that the amount of linguistic information being encoded is different across task-specific labels. In particular, language models encode more linguistic information for one label than other labels. This label-wise information difference again justifies the absence and incompleteness of some linguistic information for inference in language models. Furthermore, we found that language models can effectively learn some types of missing information on complex semantic reasoning through fine-tuning for the NLI task. Overall, language models show potential to serve as knowledge bases of linguistic information that supports robust symbolic reasoning.  

We believe that our probing and analysis provide an adequate picture of whether critical linguistic information for inference exists in pre-trained language models. Moreover, our probing can inspire future systems on combining neural and symbolic NLP methods. For future work, one could conduct further analysis on each type of linguistic information in language models by constructing more detailed probing datasets. One could also design logical systems that can access linguistic information from pre-trained language models and apply them in the inference process for improved performance on large benchmarks. 

\section{Acknowledgments}
We thank the anonymous reviewers for their thoughtful and constructive comments. Thanks also to our advisors Laurence S. Moss and Michael Wollowski for their feedback on earlier drafts of this work. Special thanks to the Machine Learning for Language Group at NYU for their wonderful NLP toolkit, JIANT \cite{phang2020jiant}.

\bibliography{aaai22}

\clearpage

\appendix
\section{Implementation Details}
We provide more implementation details on how each probing experiments is conducted, as well as the software libraries that it depends on. 
\paragraph{Software Library} For all the probing experiments, we rely on a NLP toolkit called JIANT \cite{phang2020jiant}. Jiant is a framework that supports both multi-task learning and transfer learning. We followed the edge probing experiment guide provided by the library and expanded the edge probing tasks with our own tasks. For classifiers, we used the same implementation of linear classifier and MLP classifier as the original edge probing paper \cite{edge_probing_paper}. Our pre-trained language models are from huggingface's transformers library \cite{wolf2020huggingfaces}. 

\paragraph{Hyperparameters}
We will briefly describe our selection process for key hyperparameters for all the probing tasks. For learning rate, we follow \citet{edge_probing_paper}'s practice, where 0.0001 is used for training. We performed an empirical selection for the batch size and epoch number. We set the batch size to 4 and epoch number to 10. We observed that each probing classifier requires enough training iterations and training batches to fully extract and interpret the corresponding linguistic information from the pre-trained language model representations. We also need to avoid a large amount of train data examples being exposed to the classifier to ensure that the classifier did not just learn the task. Combining these two criterion, we observe that empirically, a batch size with 4 and an epoch number with 10 is the optimal among other combinations. Overall, our selection of the important hyperparameters is summarized below:
\begin{itemize}
    \item Learning rate: $1\times10^{-4}$
    \item Batch Size: 4
    \item Epochs: 10
\end{itemize}

\paragraph{Model Freezing and Training}
For probing experiments, we follow the convention way of probing language models, by freezing the parameters of each pre-trained language model encoders. This way, the pre-trained model parameters will not update through gradient propagation and thus prevent it from learning the task. We train each probing classifiers by minimizing a loss function through gradient descent. For binary classification we use the binary cross-entropy loss. For multi-label classification, we use the softmax loss which can ensue an exclusivity constraint. For gradient descent, we use the Adam optimizer \cite{Kingma2015AdamAM} following \citet{edge_probing_paper}'s choice. 

\section{Dataset Construction}
In this section we provide more details to the construction of some datasets.
\paragraph{SemGraph} To construct the dataset used for the SemGraph task, we first generated the universal dependencies of sentences. Then we developed an algorithm to draw edges of two words as well as identifying these words' roles (i.e. concept, relation or modifier) based on the type of dependency between them. Finally we use role-labeling tools to examine the roles of vertices and refine our data.
\paragraph{Contradiction Signature}: Leveraging the fact that a pair of sentences has a similar context, we automatically mark the changing parts and then manually verified that the spans indeed contradict each other.
\paragraph{Lexical (SA-Lex)} Since two sentences in a pair of the Breaking-NLI dataset have very similar contexts, the part that is changing consists of our target spans. We automatically marked such places and verified that lexical semantic change happens from the first place to the second one. Then we randomly picked one word from each sentence to form an unaligned pair after we check that the two words are not related.
\paragraph{Anaphora (SA-AP)} Before manually annotate which entity the pronoun refers to, we utilize universal dependencies to decide the locations of potential entities and the target pronoun.
\paragraph{Sentiment (SA-ST)} We first concatenated a sample with a speech fragment from another randomly selected sample. Next we automatically draw connections (i.e. aligned or contradict) from the original speech fragment to the hypothesis based on the relation of this pair of sentences given in the original dataset. Finally, we manually checked if the added fragment could be seen as related to the hypothesis. If so, we would assign a specific label (aligned or contradict) to indicate their relation instead of classifying them as unaligned. To construct a vertex probing dataset, we assigned the \textbf{Unaligned} label to all the words outside of speech fragments and based on their relations with the hypothesis.
\paragraph{Relational Knowledge (SA-RK)} To construct this dataset, we first locate a span that contains the named entities appeared in the hypothesis. Next we manually adjust them to properly include the information needed to reveal the relation of these entities.

\end{document}